\documentclass[twocolumn,letter,10pt]{IEEEtran} 

\def \enTitle{On the Feasibility of Generic Deep \\Disaggregation for Single-Load Extraction}
\def \student{Karim Said Barsim}   
\def \worksubject{IEEE Transactions on Smart Grid}

\def \doclang{english}
\def \colortype{boxed} 	
\def \keywords
{
	Energy/Load disaggregation,
	Non-Intrusive Load Monitoring (NILM),
	Convolutional Neural Networks (CNN),
	UNet, SegNet, UK-DALE
}
\def \paperAbstract{
Recently, and with the growing development of big energy datasets, data-driven learning techniques began to represent a potential solution to the energy disaggregation problem outperforming engineered and hand-crafted models.
However, most proposed deep disaggregation models are load-dependent in the sense that either expert knowledge or a hyper-parameter optimization stage is required prior to training and deployment (normally for each load category) even upon acquisition and cleansing of aggregate and sub-metered data.
In this paper, we present a feasibility study on the development of a generic disaggregation model based on data-driven learning.
Specifically, we present a generic deep disaggregation model capable of achieving state-of-art performance in load monitoring for a variety of load categories.
The developed model is evaluated on the publicly available UK-DALE dataset with a moderately low sampling frequency and various domestic loads.
}

\usepackage[ruled,vlined]{algorithm2e}

\usepackage[cmex10]{amsmath}	

\usepackage{amssymb}

\usepackage{array}

\usepackage{booktabs}

\usepackage{bm}

\usepackage[font={footnotesize}]{caption}	

\usepackage[final]{changes}

\usepackage{dcolumn}

\usepackage{dsfont}

\usepackage{enumitem}

\usepackage{etoolbox}

\usepackage{eucal}

\usepackage[bottom]{footmisc}
\usepackage{hyperref}

\usepackage{ifthen}

\usepackage{indentfirst}

\usepackage{lipsum}

\usepackage{mathtools}

\usepackage{moresize}

\usepackage{multicol}
 
\usepackage{multirow}

\usepackage[square,numbers,sort&compress]{natbib}

\usepackage{pdfpages}

\usepackage{perpage}

\usepackage{pgfplots}

\usepackage[nodisplayskipstretch]{setspace}

\usepackage{siunitx}

\usepackage{soul}

\usepackage{stackengine}

\usepackage{stfloats}

\usepackage{subcaption}

\usepackage{tabu}

\usepackage{tabularx}

\usepackage{textcomp}

\usepackage{tikz}

    \makeatletter\newcommand\subparagraph{\@startsection{subparagraph}{5}{\parindent}{3.25ex \@plus 1ex \@minus .2ex}{-1em}{\normalfont\normalsize\bfseries}}\makeatother
\usepackage[compact]{titlesec}
    \let\subparagraph\relax

\usepackage[color = yellow]{todonotes}
 
\usepackage{url}

\usepackage{epstopdf}

\usepackage{nccmath}
\ifthenelse{\equal{\doclang}{german}}
{
	\def \langtitle{\deTitle}
		
}
{
	\def \langtitle{\enTitle}
	
}
\pdfstringdef \studentPDF {\student} 
\pdfstringdef \worktitlePDF {\langtitle}
\pdfstringdef \worksubjectPDF {\worksubject}
\pdfstringdef \keywordsPDF {\keywords}
\hypersetup{pdfauthor=\studentPDF, 
            pdftitle=\worktitlePDF,
            pdfsubject=\worksubjectPDF,
            pdfkeywords=\keywordsPDF}
\usetikzlibrary{arrows.meta,
				automata,
				backgrounds,
                calc,
				decorations,
				decorations.pathmorphing,
				decorations.pathreplacing,
                intersections,
                ocgx,
				patterns,
				petri,
				positioning,
				shapes,
				spy,
				topaths}
\pgfplotsset{compat=1.13}

\titlespacing{\section}{0pt}{3pt}{0pt}
\titlespacing{\subsection}{0pt}{3pt}{0pt}
\renewcommand{\baselinestretch}{0.90}

\MakePerPage{footnote}

\SetAlgorithmName{Algorithm}{Algorithm}{List of algorithms}

\hypersetup{pdfstartview=FitH, bookmarks=true, bookmarksnumbered=true, bookmarksopen=false, breaklinks=false,
		    citecolor = blue}
\ifthenelse{\equal{\colortype}{color}}
	{
		\hypersetup{colorlinks = true, linkcolor = blue}
	}
	{
		\ifthenelse{\equal{\colortype}{boxed}}
		{}
		{
			\hypersetup{hidelinks=true}
		}
	}

\newcommand{\on}{\emph{on}}
\newcommand{\off}{\emph{off}}
\newcommand{\plus}{\texttt{+}}
\newcommand{\minus}{\texttt{-}}

\begin{document}

\title{\enTitle}	

\author
{
\IEEEauthorblockN{Karim Said Barsim and Bin Yang\\}
\IEEEauthorblockA{\texttt{\{karim.barsim,bin.yang\}@iss.uni-stuttgart.de}\\
                    Institute of Signal Processing and System Theory, University of Stuttgart}
}

\maketitle

\begin{abstract}\paperAbstract\end{abstract}



\section{Introduction}
\label{sec:introduction}

Energy disaggregation (or \underline{N}on-\underline{I}ntrusive \underline{L}oad \underline{M}onitoring NILM) is the process of inferring individual load profiles at the end-use level from a single or a limited number of sensing points. Promising applications of disaggregated data have motivated a growing research community to reach a widely acceptable and scalable solution. Energy disaggregation  proved to be a challenging source separation problem in which a considerably large number of parameters are to be estimated from a limited set of measurements with little constraints.

In the last decade, energy disaggregation has witnessed an unprecedented wide-spreading research which is easily observed from the wide variety of learning techniques applied to this problem alongside with the growing number of energy datasets developed specifically for this research field. More recently, and analogous to the current breakthrough in data-driven learning, deep neural networks have re-gained their interest in addressing the energy disaggregation problem, especially alongside with the recently developed large energy datasets required for training such complex models \cite{Kaman_2017,Mauch_2015,Mauch_2016,Kelly_2015,Zhang_2016_SequenceToPointLearning,He_2016_AnEmpiricalStudy, Kelly_2015_UKDALE,Parson_2015_Dataport_NILMTK}. The progress in this trend, however, is relatively slow when compared to the development in either field separately. This is sometimes attributed to the high risk of over-fitting in neural network models \cite{Makonin_2014_PhD}, insufficiency or low diversity of publicly available energy datasets \cite{Kelly_2015}, or limited insights and understanding of the learning behavior of these models \cite{Zhang_2016_SequenceToPointLearning}.

In this paper, we first present a feasibility study on the development of a generic data-driven model suitable for end-use load monitoring. The proposed disaggregation model exploits a fully convolutional neural network architecture and is generic in the sense that none of the model hyper-parameters is dependent on the load category. We assess the feasibility of such a model through empirical evaluation of the monitoring performance across various load categories in a publicly available energy dataset.

\section{Related work}
\label{sec:related-work}

In this section, we briefly describe some of the most recent works on energy disaggregation and load monitoring that adopted data-driven learning techniques.

Mauch and Yang \cite{Mauch_2015} exploited a generic two-layer bidirectional \underline{R}ecurrent \underline{N}eural \underline{N}etwork (RNN) architecture featuring \underline{L}ong \underline{S}hort \underline{T}erm \underline{M}emory (LSTM) \cite{Hochreiter_1997_LSTM} units in extracting single load profiles. They tested their models on the \underline{R}efrence \underline{E}nergy \underline{D}isaggregation \underline{D}ataset (REDD) \cite{Kolter_2011_REDD} in a de-noised scheme \cite{Makonin_2015_NILMPerformanceEvaluation}. Additionally, they  validated the generalization of their architecture to previously unseen loads in new buildings. In a later work, Mauch and Yang \cite{Mauch_2016} used a combination of discriminative and generative models in a two-stage eventless extraction of load profiles. Kelly and Knottenbelt \cite{Kelly_2015} evaluated and compared three neural network architectures on domestic loads from the \underline{UK}-\underline{D}omestic \underline{A}ppliance \underline{L}evel \underline{E}nergy (UK-DALE) \cite{Kelly_2015_UKDALE}. The first is a bidirectional RNN architecture with LSTM units similar to the one in \cite{Mauch_2015}, the second follows the architecture of a \underline{d}e-noising \underline{A}uto-\underline{E}ncoder (dAE) \cite{Vincent_2010}, and the last is a regression-based disaggregator whose objective is to estimate the main key points of an activation cycle of the target load within a given window. 

Similarly, He and Chai \cite{He_2016_AnEmpiricalStudy} applied two architectures, namely a convolutional dAE and an RNN, to the same problem. In their architectures, they also applied parallel convolutional layers with different kernel sizes analogous to the Inception module in GoogLeNet \cite{Szegedy_2015}. Zhang et al. \cite{Zhang_2016_SequenceToPointLearning} simplified the objective of the dAE architecture in \cite{Kelly_2015} to predict a single time instance of the target load profile for a given window of the aggregate signal.
Likewise, Nascimento \cite{Nascimento_2016} applied three neural network architectures, namely basic convolutional dAE, an RNN, and a ResNet-based model \cite{He_2015_ResNet} to the same problem but on three target loads in the REDD dataset. He introduced several improvements such as redefining the loss function, exploiting batch normalization \cite{Ioffe_2015_BatchNormalization}, and applying residual connections \cite{He_2015_ResNet}.

Additionally, Lange et al. \cite{Lange_2016_BOLT} adopted a deep neural network with constrained binary and linear activation units in the last two layers. Their first objective was to retrieve subcomponents of the input signal that sum up linearly to the aggregate active and reactive powers. Finally, they estimate the \emph{on-off} activation vector of each load. Their approach, however, was applied on very high frequency current and voltage measurements (12 kHz) from the \underline{B}uilding-\underline{L}evel f\underline{U}lly labeled dataset for \underline{E}lectricity \underline{D}isaggregation (BLUED) \cite{Anderson_2012_BLUED}.

In many of these previous works \cite{Kelly_2015, He_2016_AnEmpiricalStudy, Zhang_2016_SequenceToPointLearning, Nascimento_2016}, each disaggregator is a neural network whose disaggregation window length (and consequently the width of subsequent layers) depends on the load being monitored. The disaggregation window of each load is manually adjusted in a per-load basis to fully capture a single activation cycle of the load. Moreover, the disaggregation performance widely differs amongst variant load categories and a model that achieves remarkably well on one load might drastically fail for other load types.

\section{Load Monitoring}
\label{sec:load-monitoring}

In this work, we focus essentially on \emph{single-load extraction of activation profiles}, of which we give a detailed description in the following.

\subsection{Activation profiles: definition and estimation}

In the simplest case, a load is modeled as a two-state machine which is assumed to be in the \on-state whenever the load is consuming energy from the main power source, and in the \off-state otherwise. Accordingly, load monitoring becomes a binary classification task.
Note that in contrast to previous works, the consumption profile of a load during its \on-state need not be a defined \cite{Zeifman_2011_VAST} nor a piecewise-defined function in time.

The desired signal (i.e. \emph{ground truth}) of a load $m$ in a window of $N$ time instances is the binary-valued signal $\underline{\omega}^{(m)}\in \{0, 1\}^N$ whose element $\omega^{(m)}(n)$ is set (i.e. to indicate an \on-state) whenever the load is operating in one of its activation states at time instance $n$ and unset otherwise. In this work, we refer to this signal as the \emph{activation profile}.
Applications that benefit from activation profiles include mainly activity monitoring and occupancy detection 
in which time-of-use information dominates the value of energy consumed.

We define the true activation profile of a load $\underline{\omega}^{(m)}$ via a threshold-based approach applied to the sub-metered real power signals and similar to the one used in \cite{Kelly_2015} as follows.
The sub-metered real-power $\underline{x}^{(m)}$ of a load $m$ is compared against predefined thresholds to detect the operation intervals of the load. In order to avoid anomalies and false activations or deactivations, the load is assumed to be in an activation state (i.e. \on) if its power draw $x^{(m)}(n)$ exceeds a given threshold $\mathcal{P}_{\text{on}}^{(m)}$ for a minimum period of time $\mathcal{N}_{\text{on}}^{(m)}$. Similarly, if the power draw drops below a predefined threshold $\mathcal{P}_{\text{off}}^{(m)}$ for a given period $\mathcal{N}_{\text{off}}^{(m)}$, the load is assumed to be disconnected. Otherwise, the load keeps its last observed state. Thus, the estimated activation profile is defined as

{\small\begin{equation*}
	\omega^{(m)}(n) = 
	\begin{cases}
	1, & \hspace{-10mm}\text{if } x^{(m)}(k) \geqslant \mathcal{P}_{\text{on}}^{(m)}, \; \text{for } n \leqslant k < n+\mathcal{N}_{\text{on}}^{(m)}\\[1mm]
	0, & \hspace{-10mm}\text{if } x^{(m)}(k) \leqslant \mathcal{P}_{\text{off}}^{(m)}, \; \text{for } n \leqslant k < n+\mathcal{N}_{\text{off}}^{(m)}\\[1mm]
	\omega^{(m)}(n-1),	& \hspace{8mm} \text{otherwise}
	\end{cases}
	\end{equation*}}%
with the initial state assumed to be \emph{off} (i.e. $\;\omega^{(m)}(0) = 0$) for all loads. Note that $\mathcal{P}^{(m)}_{\text{on}}$, $\mathcal{N}^{(m)}_{\text{on}}$, $\mathcal{P}^{(m)}_{\text{off}}$, and $\mathcal{N}^{(m)}_{\text{off}}$ are the only load-dependent parameters in this work, and they are used merely in estimating the ground truth signals. Values of these parameters are similar or close to those adopted in \cite{Kelly_2015} and are listed in Table \ref{tbl:loads} for the sake of completeness.

\subsection{Single load extraction}

In single-load extraction, each disaggregator targets exclusively a single load in the monitored circuit and normally ignores dependencies amongst loads. While exploiting loads' dependencies is expected to improve the performance of a disaggregation system in a given building \cite{Kim_2010, Kolter_2010_SparseCoding, Makonin_2014_PhD}, it is also likely to reduce the generalization capability of such a system to new, previously unseen buildings. This is because such dependencies originate not only from the physical architecture of the power line network and the assumed signal model but also from the usage behavior of end-consumers which varies widely from one building to another, especially within the residential sector \cite{Batra_2014_Comparison}.

The load monitoring problem is modeled as $K$-separate binary classification tasks. Given a window of $K$ samples of the aggregate real power signal $\underline{x}(n) = \left[x(n + k)\right]_{k=0}^{k=K-1}$ starting at the time instance $n$, the model $\underline{g}^{(m)}(\underline{x}(n), \bm{\theta})\,\in\,[0,\,1]^K$ estimates the posterior probabilities of the activation profile for the analogous $K$ time instances of $m$\textsuperscript{th} load where $\bm{\theta}$ is the model parameters (e.g. weights and biases in a neural network)

\setlength{\textfloatsep}{0.1cm}
\begin{table}[!t]
	\centering
	\footnotesize
	\caption{Load-dependent parameters for estimating the activation profiles. }
	\label{tbl:loads}
	\renewcommand{\arraystretch}{0.81}
	\footnotesize
	\begin{tabularx}{88mm}{lccc}
		\toprule
		Load  	 & $\mathcal{P}_{\text{on}} = \mathcal{P}_{\text{off}}$ [W] & $\mathcal{N}_{\text{on}}$ [min.]  & $\mathcal{N}_{\text{off}}$ [min.]\\[1mm]
		\midrule
		{Fridge \texttt{(FR)}}               & \texttt{5   }    & \texttt{1  }    & \texttt{1   }    \\[1mm]
		{Lights \texttt{(LC)}}               & \texttt{10  }    & \texttt{1  }    & \texttt{1   }    \\[1mm]
		{Dishwasher \texttt{(DW)}}           & \texttt{10  }    & \texttt{30 }    & \texttt{5   }    \\[1mm]
		{Washing machine \texttt{(WM)}}      & \texttt{20  }    & \texttt{30 }    & \texttt{5   }    \\[1mm]
		{Solar  pump \texttt{(SP)}}   		& \texttt{20  }    & \texttt{1  }    & \texttt{1   }    \\[1mm]
		{TV}                                 & \texttt{5   }    & \texttt{3  }    & \texttt{3   }    \\[1mm]
		{Boiler \texttt{(BL)}}               & \texttt{25  }    & \texttt{5  }    & \texttt{5   }    \\[1mm]
		{Kettle \texttt{(KT)}}               & \texttt{1000}    & \texttt{1/3}    & \texttt{1/6 }    \\[1mm]
		{Microwave \texttt{(MC)}}            & \texttt{50  }    & \texttt{1/6}    & \texttt{1/6 }    \\[1mm]
		{Toaster \texttt{(TS)}}              & \texttt{300 }    & \texttt{1/6}    & \texttt{1/20}    \\
		\bottomrule
	\end{tabularx}
\end{table}

{\small\begin{equation}
	p\left(\underline{\omega}^{(m)}(n)=\bm{1}\;\big|\;\underline{x}(n)\right) = \underline{g}^{(m)}\left(\underline{x}(n); \bm{\theta}\right)
	\end{equation}}%
where the disaggregator's output is bound to the valid range of a probability function via a logistic sigmoid activation in the output layer of the network

{\small\begin{equation}
	\underline{g}^{(m)}\left(\underline{x}(n); \bm{\theta}\right) = \underline{\sigma}\left(\underline{\tilde{g}}^{(m)}\left(\underline{x}(n); \bm{\theta}\right)\right)
	\end{equation}}%
where $\underline{\tilde{g}}$ represents the sub-network from the input layer to activation signals of the output layer and $\underline{\sigma}$ is the logistic sigmoid function Eq. \ref{eq:logsg-activation} applied element-wise to $\underline{\tilde{g}}$. 

In the training phase, we refer to the pair $\left(\underline{x}(n),\,\underline{\omega}(n)\right)$ as a single training segment with $K$ data samples. Training segments are extracted from the whole time series signals $\left(\underline{x},\,\underline{\omega}\right)$ using \emph{non-overlappling} windows which results in a training set whose inputs segments are

{\small\begin{equation}
	\bm{\mathcal{X}} =\left(\; \underline{x}(0),\; \underline{x}(K),\; \underline{x}(2K),\; \dots,\;\underline{x}((N_K-1)\cdot K)\right)
	\label{eq:condition-prob}
	\end{equation}}%
and the corresponding activation segments 

{\small\begin{equation}
	\bm{\Omega} = \left(\; \underline{\omega}(0),\; \underline{\omega}(K),\; \underline{\omega}(2K),\; \dots,\; \underline{\omega}((N_K-1)\cdot K)\right)
	\end{equation}}%
where $N_K = \lfloor N / K\rfloor$ is number of training segments (with the $\cdot^{(m)}$ notation omitted for brevity). Assuming all segments (and the $K$ outputs of each segment) are conditionally independent given the input vector $\underline{x}(n)$ and identically distributed (i.i.d), then the likelihood function becomes

{\small\begin{equation*}
	p\left(\bm{\Omega} |\; \bm{\mathcal{X}}, \bm{\theta} \right) = \prod_{n=0}^{N_K-1} \prod_{k=0}^{K-1} p\left(\omega(k + n\cdot K) \;|\; \underline{x}(n\cdot K)\right)
	\end{equation*}}%
with $\omega$ being a Bernoulli distributed random variable

{\small\begin{equation*}
	p\left(\bm{\Omega} |\; \bm{\mathcal{X}}, \bm{\theta} \right) = \prod_{n=0}^{N_K-1} \prod_{k=0}^{K-1} g_k(\underline{x}(n))^{\omega_k} \cdot (1 - g_k(\underline{x}(n)))^{1-\omega_k}
	\end{equation*}}%
where $g_k(\underline{x}(n))$ is the $k^\text{th}$ output of the disaggregator. The negative log-likelihood $\texttt{\textbf{-}}\mathcal{LL}$ then becomes

{\small\begin{equation*}
	\small
	\texttt{\textbf{-}}\mathcal{LL} = -\kern-0.5em\sum_{n=0}^{N_K-1} \sum_{k=0}^{K-1} \omega_k\cdot\ln g_k(\underline{x}(n)) + (1-\omega_k) \ln (1 - g_k(\underline{x}(n)))
	\end{equation*}}%
which is known as \emph{binary cross-entropy} and it is the adopted loss function in all our experiments. The choice of a logistic sigmoid activation function in the output layer together with the binary cross-entropy as the objective function is a standard combination in binary classification problems \cite{Bishop_2006}.

Finally, the following decision rule is used to estimate the final class labels

{\small\begin{equation*}
	\hat{\omega}(n)=
	\begin{cases}
	0 & \text{if } p(\omega(n) = 0\,|\,\underline{x}(n)) > p(\omega(n) = 1\,|\,\underline{x}(n)) \\[1mm]
	1 & \text{otherwise}
	\end{cases}
	\end{equation*}}%

We point out that the concept of load activation cycles, a complete cycle of operation \off $\rightarrow$ \on $\rightarrow$ \off, is not considered. In other words, an activation cycle of a load can extend over several $K$-length segments (such as lighting circuits and dishwashers) or arise more than once within the same window segment (as in fridge and kettle activations). This is an important property since a disaggregator need not wait till the deactivation of a load (i.e. switch-\off event) but rather can provide {near} real-time feedback from a partial segment of the activation, normally with some delay. 

\section{Model Architecture}
\label{sec:model-selection}

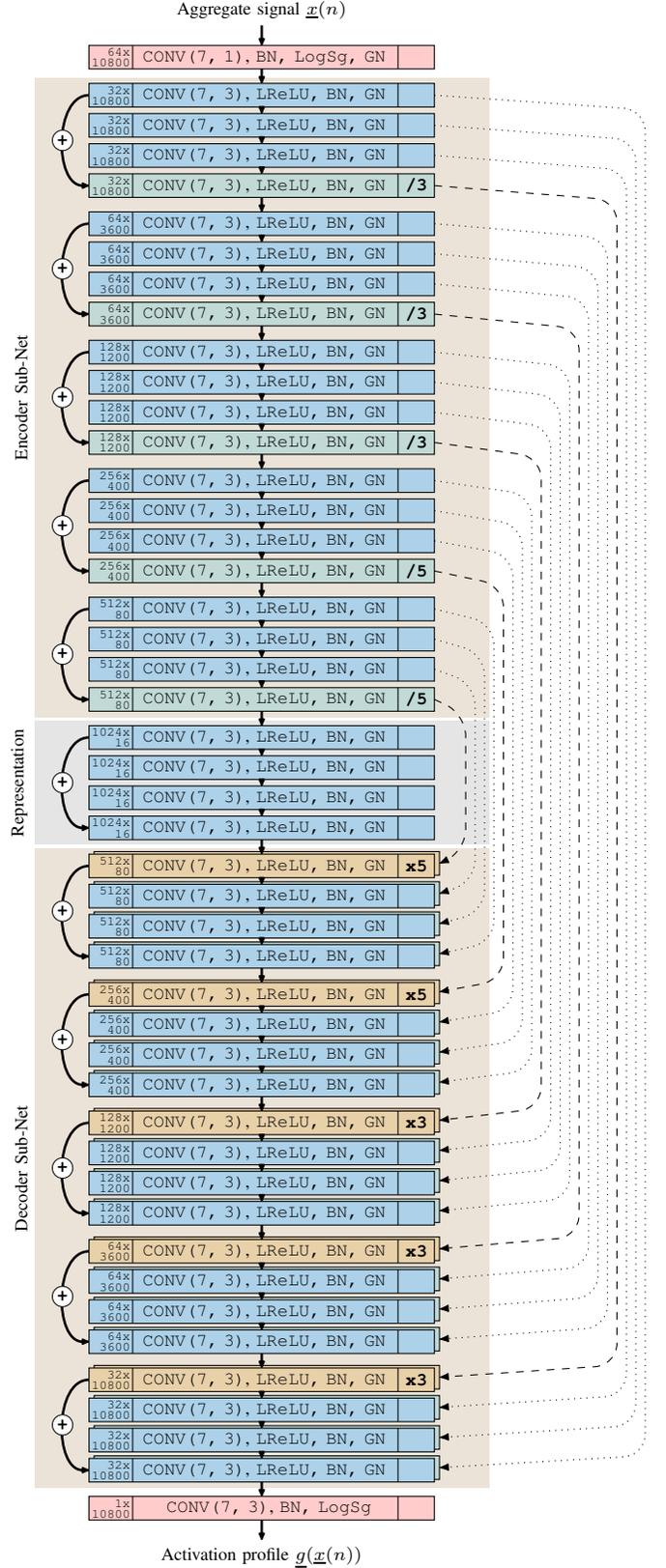
\begin{figure}[!htbp]
	\definecolor{bc01}{RGB}{232, 208, 169}
	\definecolor{bc02}{RGB}{183, 175, 163}
	\definecolor{bc03}{RGB}{193, 218, 214}
	\definecolor{bc04}{RGB}{172, 209, 233}
	\definecolor{bc05}{RGB}{109, 146, 155}
	\definecolor{bc06}{RGB}{245, 250, 250}
	\definecolor{bsgm}{RGB}{255, 204, 203}
	\definecolor{mainblocknormal}{RGB}{0, 153, 204}
	\definecolor{pooling}{RGB}{67,171,101}
	\definecolor{upsampling}{RGB}{81, 163, 157}
	\pgfmathsetlengthmacro\LblR{5mm}
	\pgfmathsetlengthmacro\LblL{6mm}
	\pgfmathsetlengthmacro\AllLayerWidth{1.75mm}
	\pgfmathsetlengthmacro\InnerLayerSep{3.9mm}
	\pgfmathsetlengthmacro\InnerLayerSepMerge{0.6mm}
	\pgfmathsetlengthmacro\SkpXShift{2mm}	
	
	\pgfdeclarelayer{bbackground}
	\pgfdeclarelayer{background}
	\pgfdeclarelayer{foreground}
	\pgfsetlayers{bbackground,background,main,foreground}

	\def\vsvsin{-2.75mm}
	\def\vshdin{-0.90mm}
	\def\vshdot{-1.99mm}
	\def\vsvsot{-2.75mm}
	
	\def\CONVHeight{3.25mm}
	\def\CONVWidth{33mm}
	
	\tikzstyle{CONVst}=[anchor=north, draw=black, minimum height=\CONVHeight, minimum width=\CONVWidth,
	inner xsep=0mm, inner ysep=0mm, outer xsep=0mm, outer ysep=0mm, text width=47.5mm]
	\tikzstyle{llst}=[anchor=east, align=right, inner xsep=0.3mm, inner ysep=0mm, font={\tiny}, align=right]
	\tikzstyle{rlst}=[anchor=west, align=right, inner xsep=1.0mm, inner ysep=0mm, font={\scriptsize}]
	\tikzstyle{sqar}=[line width=1.0pt, arrows={-{Latex[length=1.5mm, round]}}]
	\tikzstyle{skrs}=[]
	\tikzstyle{idad}=[]

	\centering
	\scriptsize
	\begin{tikzpicture}

	\node (inpt) [anchor=north] {Aggregate signal $\underline{x}(n)$};
	\node (x000) at ($(inpt.south) + (0, \vsvsin)$) [CONVst, fill=bsgm] {\texttt{\ \ \ \ \ \ \ \ \ CONV(7, 1)}, \texttt{BN, LogSg, GN}};↕ \node (ll00) at ($(x000.west) + (\LblL, 0)$) [llst] {\texttt{  64x}\\[-0.6mm]\texttt{10800}}; \node (rl00) at ($(x000.east) - (\LblR, 0)$) [rlst] {\texttt{\textbf{  }}};
	\node (xl10) at ($(x000.south) + (0, \vshdot)$) [CONVst, fill=bc04] {\texttt{\ \ \ \ \ \ \ \ \ CONV(7, 3)}, \texttt{LReLU, BN, GN}};↕ \node (ll10) at ($(xl10.west) + (\LblL, 0)$) [llst] {\texttt{  32x}\\[-0.6mm]\texttt{10800}}; \node (rl10) at ($(xl10.east) - (\LblR, 0)$) [rlst] {\texttt{\textbf{  }}};  
	\node (xl11) at ($(xl10.south) + (0, \vshdin)$) [CONVst, fill=bc04] {\texttt{\ \ \ \ \ \ \ \ \ CONV(7, 3)}, \texttt{LReLU, BN, GN}};↕ \node (ll11) at ($(xl11.west) + (\LblL, 0)$) [llst] {\texttt{  32x}\\[-0.6mm]\texttt{10800}}; \node (rl11) at ($(xl11.east) - (\LblR, 0)$) [rlst] {\texttt{\textbf{  }}};  
	\node (xl12) at ($(xl11.south) + (0, \vshdin)$) [CONVst, fill=bc04] {\texttt{\ \ \ \ \ \ \ \ \ CONV(7, 3)}, \texttt{LReLU, BN, GN}};↕ \node (ll12) at ($(xl12.west) + (\LblL, 0)$) [llst] {\texttt{  32x}\\[-0.6mm]\texttt{10800}}; \node (rl12) at ($(xl12.east) - (\LblR, 0)$) [rlst] {\texttt{\textbf{  }}};  
	\node (xl13) at ($(xl12.south) + (0, \vshdin)$) [CONVst, fill=bc03] {\texttt{\ \ \ \ \ \ \ \ \ CONV(7, 3)}, \texttt{LReLU, BN, GN}};↕ \node (ll13) at ($(xl13.west) + (\LblL, 0)$) [llst] {\texttt{  32x}\\[-0.6mm]\texttt{10800}}; \node (rl13) at ($(xl13.east) - (\LblR, 0)$) [rlst] {\texttt{\textbf{/3}}};  
	\node (xl20) at ($(xl13.south) + (0, \vshdot)$) [CONVst, fill=bc04] {\texttt{\ \ \ \ \ \ \ \ \ CONV(7, 3)}, \texttt{LReLU, BN, GN}};↕ \node (ll20) at ($(xl20.west) + (\LblL, 0)$) [llst] {\texttt{  64x}\\[-0.6mm]\texttt{ 3600}}; \node (rl20) at ($(xl20.east) - (\LblR, 0)$) [rlst] {\texttt{\textbf{  }}};  
	\node (xl21) at ($(xl20.south) + (0, \vshdin)$) [CONVst, fill=bc04] {\texttt{\ \ \ \ \ \ \ \ \ CONV(7, 3)}, \texttt{LReLU, BN, GN}};↕ \node (ll21) at ($(xl21.west) + (\LblL, 0)$) [llst] {\texttt{  64x}\\[-0.6mm]\texttt{ 3600}}; \node (rl21) at ($(xl21.east) - (\LblR, 0)$) [rlst] {\texttt{\textbf{  }}};  
	\node (xl22) at ($(xl21.south) + (0, \vshdin)$) [CONVst, fill=bc04] {\texttt{\ \ \ \ \ \ \ \ \ CONV(7, 3)}, \texttt{LReLU, BN, GN}};↕ \node (ll22) at ($(xl22.west) + (\LblL, 0)$) [llst] {\texttt{  64x}\\[-0.6mm]\texttt{ 3600}}; \node (rl22) at ($(xl22.east) - (\LblR, 0)$) [rlst] {\texttt{\textbf{  }}};  
	\node (xl23) at ($(xl22.south) + (0, \vshdin)$) [CONVst, fill=bc03] {\texttt{\ \ \ \ \ \ \ \ \ CONV(7, 3)}, \texttt{LReLU, BN, GN}};↕ \node (ll23) at ($(xl23.west) + (\LblL, 0)$) [llst] {\texttt{  64x}\\[-0.6mm]\texttt{ 3600}}; \node (rl23) at ($(xl23.east) - (\LblR, 0)$) [rlst] {\texttt{\textbf{/3}}};  
	\node (xl30) at ($(xl23.south) + (0, \vshdot)$) [CONVst, fill=bc04] {\texttt{\ \ \ \ \ \ \ \ \ CONV(7, 3)}, \texttt{LReLU, BN, GN}};↕ \node (ll30) at ($(xl30.west) + (\LblL, 0)$) [llst] {\texttt{ 128x}\\[-0.6mm]\texttt{ 1200}}; \node (rl30) at ($(xl30.east) - (\LblR, 0)$) [rlst] {\texttt{\textbf{  }}};  
	\node (xl31) at ($(xl30.south) + (0, \vshdin)$) [CONVst, fill=bc04] {\texttt{\ \ \ \ \ \ \ \ \ CONV(7, 3)}, \texttt{LReLU, BN, GN}};↕ \node (ll31) at ($(xl31.west) + (\LblL, 0)$) [llst] {\texttt{ 128x}\\[-0.6mm]\texttt{ 1200}}; \node (rl31) at ($(xl31.east) - (\LblR, 0)$) [rlst] {\texttt{\textbf{  }}};  
	\node (xl32) at ($(xl31.south) + (0, \vshdin)$) [CONVst, fill=bc04] {\texttt{\ \ \ \ \ \ \ \ \ CONV(7, 3)}, \texttt{LReLU, BN, GN}};↕ \node (ll32) at ($(xl32.west) + (\LblL, 0)$) [llst] {\texttt{ 128x}\\[-0.6mm]\texttt{ 1200}}; \node (rl32) at ($(xl32.east) - (\LblR, 0)$) [rlst] {\texttt{\textbf{  }}};  
	\node (xl33) at ($(xl32.south) + (0, \vshdin)$) [CONVst, fill=bc03] {\texttt{\ \ \ \ \ \ \ \ \ CONV(7, 3)}, \texttt{LReLU, BN, GN}};↕ \node (ll33) at ($(xl33.west) + (\LblL, 0)$) [llst] {\texttt{ 128x}\\[-0.6mm]\texttt{ 1200}}; \node (rl33) at ($(xl33.east) - (\LblR, 0)$) [rlst] {\texttt{\textbf{/3}}};  
	\node (xl40) at ($(xl33.south) + (0, \vshdot)$) [CONVst, fill=bc04] {\texttt{\ \ \ \ \ \ \ \ \ CONV(7, 3)}, \texttt{LReLU, BN, GN}};↕ \node (ll40) at ($(xl40.west) + (\LblL, 0)$) [llst] {\texttt{ 256x}\\[-0.6mm]\texttt{  400}}; \node (rl40) at ($(xl40.east) - (\LblR, 0)$) [rlst] {\texttt{\textbf{  }}}; 
	\node (xl41) at ($(xl40.south) + (0, \vshdin)$) [CONVst, fill=bc04] {\texttt{\ \ \ \ \ \ \ \ \ CONV(7, 3)}, \texttt{LReLU, BN, GN}};↕ \node (ll41) at ($(xl41.west) + (\LblL, 0)$) [llst] {\texttt{ 256x}\\[-0.6mm]\texttt{  400}}; \node (rl41) at ($(xl41.east) - (\LblR, 0)$) [rlst] {\texttt{\textbf{  }}};  
	\node (xl42) at ($(xl41.south) + (0, \vshdin)$) [CONVst, fill=bc04] {\texttt{\ \ \ \ \ \ \ \ \ CONV(7, 3)}, \texttt{LReLU, BN, GN}};↕ \node (ll42) at ($(xl42.west) + (\LblL, 0)$) [llst] {\texttt{ 256x}\\[-0.6mm]\texttt{  400}}; \node (rl42) at ($(xl42.east) - (\LblR, 0)$) [rlst] {\texttt{\textbf{  }}};  
	\node (xl43) at ($(xl42.south) + (0, \vshdin)$) [CONVst, fill=bc03] {\texttt{\ \ \ \ \ \ \ \ \ CONV(7, 3)}, \texttt{LReLU, BN, GN}};↕ \node (ll43) at ($(xl43.west) + (\LblL, 0)$) [llst] {\texttt{ 256x}\\[-0.6mm]\texttt{  400}}; \node (rl43) at ($(xl43.east) - (\LblR, 0)$) [rlst] {\texttt{\textbf{/5}}};  
	\node (xl50) at ($(xl43.south) + (0, \vshdot)$) [CONVst, fill=bc04] {\texttt{\ \ \ \ \ \ \ \ \ CONV(7, 3)}, \texttt{LReLU, BN, GN}};↕ \node (ll50) at ($(xl50.west) + (\LblL, 0)$) [llst] {\texttt{ 512x}\\[-0.6mm]\texttt{   80}}; \node (rl50) at ($(xl50.east) - (\LblR, 0)$) [rlst] {\texttt{\textbf{  }}};  
	\node (xl51) at ($(xl50.south) + (0, \vshdin)$) [CONVst, fill=bc04] {\texttt{\ \ \ \ \ \ \ \ \ CONV(7, 3)}, \texttt{LReLU, BN, GN}};↕ \node (ll51) at ($(xl51.west) + (\LblL, 0)$) [llst] {\texttt{ 512x}\\[-0.6mm]\texttt{   80}}; \node (rl51) at ($(xl51.east) - (\LblR, 0)$) [rlst] {\texttt{\textbf{  }}};  
	\node (xl52) at ($(xl51.south) + (0, \vshdin)$) [CONVst, fill=bc04] {\texttt{\ \ \ \ \ \ \ \ \ CONV(7, 3)}, \texttt{LReLU, BN, GN}};↕ \node (ll52) at ($(xl52.west) + (\LblL, 0)$) [llst] {\texttt{ 512x}\\[-0.6mm]\texttt{   80}}; \node (rl52) at ($(xl52.east) - (\LblR, 0)$) [rlst] {\texttt{\textbf{  }}};  
	\node (xl53) at ($(xl52.south) + (0, \vshdin)$) [CONVst, fill=bc03] {\texttt{\ \ \ \ \ \ \ \ \ CONV(7, 3)}, \texttt{LReLU, BN, GN}};↕ \node (ll53) at ($(xl53.west) + (\LblL, 0)$) [llst] {\texttt{ 512x}\\[-0.6mm]\texttt{   80}}; \node (rl53) at ($(xl53.east) - (\LblR, 0)$) [rlst] {\texttt{\textbf{/5}}};  
	\node (xl60) at ($(xl53.south) + (0, \vshdot)$) [CONVst, fill=bc04] {\texttt{\ \ \ \ \ \ \ \ \ CONV(7, 3)}, \texttt{LReLU, BN, GN}};↕ \node (ll60) at ($(xl60.west) + (\LblL, 0)$) [llst] {\texttt{1024x}\\[-0.6mm]\texttt{   16}}; \node (rl60) at ($(xl60.east) - (\LblR, 0)$) [rlst] {\texttt{\textbf{  }}};  
	\node (xl61) at ($(xl60.south) + (0, \vshdin)$) [CONVst, fill=bc04] {\texttt{\ \ \ \ \ \ \ \ \ CONV(7, 3)}, \texttt{LReLU, BN, GN}};↕ \node (ll61) at ($(xl61.west) + (\LblL, 0)$) [llst] {\texttt{1024x}\\[-0.6mm]\texttt{   16}}; \node (rl61) at ($(xl61.east) - (\LblR, 0)$) [rlst] {\texttt{\textbf{  }}};  
	\node (xl62) at ($(xl61.south) + (0, \vshdin)$) [CONVst, fill=bc04] {\texttt{\ \ \ \ \ \ \ \ \ CONV(7, 3)}, \texttt{LReLU, BN, GN}};↕ \node (ll62) at ($(xl62.west) + (\LblL, 0)$) [llst] {\texttt{1024x}\\[-0.6mm]\texttt{   16}}; \node (rl62) at ($(xl62.east) - (\LblR, 0)$) [rlst] {\texttt{\textbf{  }}};  
	\node (xl63) at ($(xl62.south) + (0, \vshdin)$) [CONVst, fill=bc04] {\texttt{\ \ \ \ \ \ \ \ \ CONV(7, 3)}, \texttt{LReLU, BN, GN}};↕ \node (ll63) at ($(xl63.west) + (\LblL, 0)$) [llst] {\texttt{1024x}\\[-0.6mm]\texttt{   16}}; \node (rl63) at ($(xl63.east) - (\LblR, 0)$) [rlst] {\texttt{\textbf{  }}};  
	\node (xl70) at ($(xl63.south) + (0, \vshdot)$) [CONVst, fill=bc01] {\texttt{\ \ \ \ \ \ \ \ \ CONV(7, 3)}, \texttt{LReLU, BN, GN}};↕ \node (ll70) at ($(xl70.west) + (\LblL, 0)$) [llst] {\texttt{ 512x}\\[-0.6mm]\texttt{   80}}; \node (rl70) at ($(xl70.east) - (\LblR, 0)$) [rlst] {\texttt{\textbf{x5}}};  
	\node (xl71) at ($(xl70.south) + (0, \vshdin)$) [CONVst, fill=bc04] {\texttt{\ \ \ \ \ \ \ \ \ CONV(7, 3)}, \texttt{LReLU, BN, GN}};↕ \node (ll71) at ($(xl71.west) + (\LblL, 0)$) [llst] {\texttt{ 512x}\\[-0.6mm]\texttt{   80}}; \node (rl71) at ($(xl71.east) - (\LblR, 0)$) [rlst] {\texttt{\textbf{  }}};  
	\node (xl72) at ($(xl71.south) + (0, \vshdin)$) [CONVst, fill=bc04] {\texttt{\ \ \ \ \ \ \ \ \ CONV(7, 3)}, \texttt{LReLU, BN, GN}};↕ \node (ll72) at ($(xl72.west) + (\LblL, 0)$) [llst] {\texttt{ 512x}\\[-0.6mm]\texttt{   80}}; \node (rl72) at ($(xl72.east) - (\LblR, 0)$) [rlst] {\texttt{\textbf{  }}};  
	\node (xl73) at ($(xl72.south) + (0, \vshdin)$) [CONVst, fill=bc04] {\texttt{\ \ \ \ \ \ \ \ \ CONV(7, 3)}, \texttt{LReLU, BN, GN}};↕ \node (ll73) at ($(xl73.west) + (\LblL, 0)$) [llst] {\texttt{ 512x}\\[-0.6mm]\texttt{   80}}; \node (rl73) at ($(xl73.east) - (\LblR, 0)$) [rlst] {\texttt{\textbf{  }}};  
	\node (xl80) at ($(xl73.south) + (0, \vshdot)$) [CONVst, fill=bc01] {\texttt{\ \ \ \ \ \ \ \ \ CONV(7, 3)}, \texttt{LReLU, BN, GN}};↕ \node (ll80) at ($(xl80.west) + (\LblL, 0)$) [llst] {\texttt{ 256x}\\[-0.6mm]\texttt{  400}}; \node (rl80) at ($(xl80.east) - (\LblR, 0)$) [rlst] {\texttt{\textbf{x5}}};  
	\node (xl81) at ($(xl80.south) + (0, \vshdin)$) [CONVst, fill=bc04] {\texttt{\ \ \ \ \ \ \ \ \ CONV(7, 3)}, \texttt{LReLU, BN, GN}};↕ \node (ll81) at ($(xl81.west) + (\LblL, 0)$) [llst] {\texttt{ 256x}\\[-0.6mm]\texttt{  400}}; \node (rl81) at ($(xl81.east) - (\LblR, 0)$) [rlst] {\texttt{\textbf{  }}};  
	\node (xl82) at ($(xl81.south) + (0, \vshdin)$) [CONVst, fill=bc04] {\texttt{\ \ \ \ \ \ \ \ \ CONV(7, 3)}, \texttt{LReLU, BN, GN}};↕ \node (ll82) at ($(xl82.west) + (\LblL, 0)$) [llst] {\texttt{ 256x}\\[-0.6mm]\texttt{  400}}; \node (rl82) at ($(xl82.east) - (\LblR, 0)$) [rlst] {\texttt{\textbf{  }}};  
	\node (xl83) at ($(xl82.south) + (0, \vshdin)$) [CONVst, fill=bc04] {\texttt{\ \ \ \ \ \ \ \ \ CONV(7, 3)}, \texttt{LReLU, BN, GN}};↕ \node (ll83) at ($(xl83.west) + (\LblL, 0)$) [llst] {\texttt{ 256x}\\[-0.6mm]\texttt{  400}}; \node (rl83) at ($(xl83.east) - (\LblR, 0)$) [rlst] {\texttt{\textbf{  }}};  
	\node (xl90) at ($(xl83.south) + (0, \vshdot)$) [CONVst, fill=bc01] {\texttt{\ \ \ \ \ \ \ \ \ CONV(7, 3)}, \texttt{LReLU, BN, GN}};↕ \node (ll90) at ($(xl90.west) + (\LblL, 0)$) [llst] {\texttt{ 128x}\\[-0.6mm]\texttt{ 1200}}; \node (rl90) at ($(xl90.east) - (\LblR, 0)$) [rlst] {\texttt{\textbf{x3}}};  
	\node (xl91) at ($(xl90.south) + (0, \vshdin)$) [CONVst, fill=bc04] {\texttt{\ \ \ \ \ \ \ \ \ CONV(7, 3)}, \texttt{LReLU, BN, GN}};↕ \node (ll91) at ($(xl91.west) + (\LblL, 0)$) [llst] {\texttt{ 128x}\\[-0.6mm]\texttt{ 1200}}; \node (rl91) at ($(xl91.east) - (\LblR, 0)$) [rlst] {\texttt{\textbf{  }}};  
	\node (xl92) at ($(xl91.south) + (0, \vshdin)$) [CONVst, fill=bc04] {\texttt{\ \ \ \ \ \ \ \ \ CONV(7, 3)}, \texttt{LReLU, BN, GN}};↕ \node (ll92) at ($(xl92.west) + (\LblL, 0)$) [llst] {\texttt{ 128x}\\[-0.6mm]\texttt{ 1200}}; \node (rl92) at ($(xl92.east) - (\LblR, 0)$) [rlst] {\texttt{\textbf{  }}};  
	\node (xl93) at ($(xl92.south) + (0, \vshdin)$) [CONVst, fill=bc04] {\texttt{\ \ \ \ \ \ \ \ \ CONV(7, 3)}, \texttt{LReLU, BN, GN}};↕ \node (ll93) at ($(xl93.west) + (\LblL, 0)$) [llst] {\texttt{ 128x}\\[-0.6mm]\texttt{ 1200}}; \node (rl93) at ($(xl93.east) - (\LblR, 0)$) [rlst] {\texttt{\textbf{  }}};  
	\node (xlA0) at ($(xl93.south) + (0, \vshdot)$) [CONVst, fill=bc01] {\texttt{\ \ \ \ \ \ \ \ \ CONV(7, 3)}, \texttt{LReLU, BN, GN}};↕ \node (llA0) at ($(xlA0.west) + (\LblL, 0)$) [llst] {\texttt{  64x}\\[-0.6mm]\texttt{ 3600}}; \node (rlA0) at ($(xlA0.east) - (\LblR, 0)$) [rlst] {\texttt{\textbf{x3}}};  
	\node (xlA1) at ($(xlA0.south) + (0, \vshdin)$) [CONVst, fill=bc04] {\texttt{\ \ \ \ \ \ \ \ \ CONV(7, 3)}, \texttt{LReLU, BN, GN}};↕ \node (llA1) at ($(xlA1.west) + (\LblL, 0)$) [llst] {\texttt{  64x}\\[-0.6mm]\texttt{ 3600}}; \node (rlA1) at ($(xlA1.east) - (\LblR, 0)$) [rlst] {\texttt{\textbf{  }}};  
	\node (xlA2) at ($(xlA1.south) + (0, \vshdin)$) [CONVst, fill=bc04] {\texttt{\ \ \ \ \ \ \ \ \ CONV(7, 3)}, \texttt{LReLU, BN, GN}};↕ \node (llA2) at ($(xlA2.west) + (\LblL, 0)$) [llst] {\texttt{  64x}\\[-0.6mm]\texttt{ 3600}}; \node (rlA2) at ($(xlA2.east) - (\LblR, 0)$) [rlst] {\texttt{\textbf{  }}};  
	\node (xlA3) at ($(xlA2.south) + (0, \vshdin)$) [CONVst, fill=bc04] {\texttt{\ \ \ \ \ \ \ \ \ CONV(7, 3)}, \texttt{LReLU, BN, GN}};↕ \node (llA3) at ($(xlA3.west) + (\LblL, 0)$) [llst] {\texttt{  64x}\\[-0.6mm]\texttt{ 3600}}; \node (rlA3) at ($(xlA3.east) - (\LblR, 0)$) [rlst] {\texttt{\textbf{  }}};  
	\node (xlB0) at ($(xlA3.south) + (0, \vshdot)$) [CONVst, fill=bc01] {\texttt{\ \ \ \ \ \ \ \ \ CONV(7, 3)}, \texttt{LReLU, BN, GN}};↕ \node (llB0) at ($(xlB0.west) + (\LblL, 0)$) [llst] {\texttt{  32x}\\[-0.6mm]\texttt{10800}}; \node (rlB0) at ($(xlB0.east) - (\LblR, 0)$) [rlst] {\texttt{\textbf{x3}}};  
	\node (xlB1) at ($(xlB0.south) + (0, \vshdin)$) [CONVst, fill=bc04] {\texttt{\ \ \ \ \ \ \ \ \ CONV(7, 3)}, \texttt{LReLU, BN, GN}};↕ \node (llB1) at ($(xlB1.west) + (\LblL, 0)$) [llst] {\texttt{  32x}\\[-0.6mm]\texttt{10800}}; \node (rlB1) at ($(xlB1.east) - (\LblR, 0)$) [rlst] {\texttt{\textbf{  }}};  
	\node (xlB2) at ($(xlB1.south) + (0, \vshdin)$) [CONVst, fill=bc04] {\texttt{\ \ \ \ \ \ \ \ \ CONV(7, 3)}, \texttt{LReLU, BN, GN}};↕ \node (llB2) at ($(xlB2.west) + (\LblL, 0)$) [llst] {\texttt{  32x}\\[-0.6mm]\texttt{10800}}; \node (rlB2) at ($(xlB2.east) - (\LblR, 0)$) [rlst] {\texttt{\textbf{  }}};  
	\node (xlB3) at ($(xlB2.south) + (0, \vshdin)$) [CONVst, fill=bc04] {\texttt{\ \ \ \ \ \ \ \ \ CONV(7, 3)}, \texttt{LReLU, BN, GN}};↕ \node (llB3) at ($(xlB3.west) + (\LblL, 0)$) [llst] {\texttt{  32x}\\[-0.6mm]\texttt{10800}}; \node (rlB3) at ($(xlB3.east) - (\LblR, 0)$) [rlst] {\texttt{\textbf{  }}};  
	\node (xlD0) at ($(xlB3.south) + (0, \vshdot)$) [CONVst, fill=bsgm] {\texttt{\ \ \ \ \ \ \ \ \ \ \ \ \ CONV(7, 3)}, \texttt{BN, LogSg    }};↕ \node (llD0) at ($(xlD0.west) + (\LblL, 0)$) [llst] {\texttt{   1x}\\[-0.6mm]\texttt{10800}}; \node (rlD0) at ($(xlD0.east) - (\LblR, 0)$) [rlst] {\texttt{\textbf{  }}};
	\node (otpt) at ($(xlD0.south) + (0, \vsvsot)$) [anchor=north]  {Activation profile $\underline{g}(\underline{x}(n))$};

	\draw (x000.north -| rl00.west) -- (x000.south -| rl00.west) (x000.north -| llD0.east) -- (x000.south -| llD0.east);
	\draw (xl10.north -| rl10.west) -- (xl10.south -| rl10.west) (xl10.north -| ll00.east) -- (xl10.south -| ll00.east);
	\draw (xl11.north -| rl11.west) -- (xl11.south -| rl11.west) (xl11.north -| ll10.east) -- (xl11.south -| ll10.east);
	\draw (xl12.north -| rl12.west) -- (xl12.south -| rl12.west) (xl12.north -| ll11.east) -- (xl12.south -| ll11.east);
	\draw (xl13.north -| rl13.west) -- (xl13.south -| rl13.west) (xl13.north -| ll12.east) -- (xl13.south -| ll12.east);
	\draw (xl20.north -| rl20.west) -- (xl20.south -| rl20.west) (xl20.north -| ll13.east) -- (xl20.south -| ll13.east);
	\draw (xl21.north -| rl21.west) -- (xl21.south -| rl21.west) (xl21.north -| ll20.east) -- (xl21.south -| ll20.east);
	\draw (xl22.north -| rl22.west) -- (xl22.south -| rl22.west) (xl22.north -| ll21.east) -- (xl22.south -| ll21.east);
	\draw (xl23.north -| rl23.west) -- (xl23.south -| rl23.west) (xl23.north -| ll22.east) -- (xl23.south -| ll22.east);
	\draw (xl30.north -| rl30.west) -- (xl30.south -| rl30.west) (xl30.north -| ll23.east) -- (xl30.south -| ll23.east);
	\draw (xl31.north -| rl31.west) -- (xl31.south -| rl31.west) (xl31.north -| ll30.east) -- (xl31.south -| ll30.east);
	\draw (xl32.north -| rl32.west) -- (xl32.south -| rl32.west) (xl32.north -| ll31.east) -- (xl32.south -| ll31.east);
	\draw (xl33.north -| rl33.west) -- (xl33.south -| rl33.west) (xl33.north -| ll32.east) -- (xl33.south -| ll32.east);
	\draw (xl40.north -| rl40.west) -- (xl40.south -| rl40.west) (xl40.north -| ll33.east) -- (xl40.south -| ll33.east);
	\draw (xl41.north -| rl41.west) -- (xl41.south -| rl41.west) (xl41.north -| ll40.east) -- (xl41.south -| ll40.east);
	\draw (xl42.north -| rl42.west) -- (xl42.south -| rl42.west) (xl42.north -| ll41.east) -- (xl42.south -| ll41.east);
	\draw (xl43.north -| rl43.west) -- (xl43.south -| rl43.west) (xl43.north -| ll42.east) -- (xl43.south -| ll42.east);
	\draw (xl50.north -| rl50.west) -- (xl50.south -| rl50.west) (xl50.north -| ll43.east) -- (xl50.south -| ll43.east);
	\draw (xl51.north -| rl51.west) -- (xl51.south -| rl51.west) (xl51.north -| ll50.east) -- (xl51.south -| ll50.east);
	\draw (xl52.north -| rl52.west) -- (xl52.south -| rl52.west) (xl52.north -| ll51.east) -- (xl52.south -| ll51.east);
	\draw (xl53.north -| rl53.west) -- (xl53.south -| rl53.west) (xl53.north -| ll52.east) -- (xl53.south -| ll52.east);
	\draw (xl60.north -| rl60.west) -- (xl60.south -| rl60.west) (xl60.north -| ll53.east) -- (xl60.south -| ll53.east);
	\draw (xl61.north -| rl61.west) -- (xl61.south -| rl61.west) (xl61.north -| ll60.east) -- (xl61.south -| ll60.east);
	\draw (xl62.north -| rl62.west) -- (xl62.south -| rl62.west) (xl62.north -| ll61.east) -- (xl62.south -| ll61.east);
	\draw (xl63.north -| rl63.west) -- (xl63.south -| rl63.west) (xl63.north -| ll62.east) -- (xl63.south -| ll62.east);
	\draw (xl70.north -| rl70.west) -- (xl70.south -| rl70.west) (xl70.north -| ll63.east) -- (xl70.south -| ll63.east);
	\draw (xl71.north -| rl71.west) -- (xl71.south -| rl71.west) (xl71.north -| ll70.east) -- (xl71.south -| ll70.east);
	\draw (xl72.north -| rl72.west) -- (xl72.south -| rl72.west) (xl72.north -| ll71.east) -- (xl72.south -| ll71.east);
	\draw (xl73.north -| rl73.west) -- (xl73.south -| rl73.west) (xl73.north -| ll72.east) -- (xl73.south -| ll72.east);
	\draw (xl80.north -| rl80.west) -- (xl80.south -| rl80.west) (xl80.north -| ll73.east) -- (xl80.south -| ll73.east);
	\draw (xl81.north -| rl81.west) -- (xl81.south -| rl81.west) (xl81.north -| ll80.east) -- (xl81.south -| ll80.east);
	\draw (xl82.north -| rl82.west) -- (xl82.south -| rl82.west) (xl82.north -| ll81.east) -- (xl82.south -| ll81.east);
	\draw (xl83.north -| rl83.west) -- (xl83.south -| rl83.west) (xl83.north -| ll82.east) -- (xl83.south -| ll82.east);
	\draw (xl90.north -| rl90.west) -- (xl90.south -| rl90.west) (xl90.north -| ll83.east) -- (xl90.south -| ll83.east);
	\draw (xl91.north -| rl91.west) -- (xl91.south -| rl91.west) (xl91.north -| ll90.east) -- (xl91.south -| ll90.east);
	\draw (xl92.north -| rl92.west) -- (xl92.south -| rl92.west) (xl92.north -| ll91.east) -- (xl92.south -| ll91.east);
	\draw (xl93.north -| rl93.west) -- (xl93.south -| rl93.west) (xl93.north -| ll92.east) -- (xl93.south -| ll92.east);
	\draw (xlA0.north -| rlA0.west) -- (xlA0.south -| rlA0.west) (xlA0.north -| ll93.east) -- (xlA0.south -| ll93.east);
	\draw (xlA1.north -| rlA1.west) -- (xlA1.south -| rlA1.west) (xlA1.north -| llA0.east) -- (xlA1.south -| llA0.east);
	\draw (xlA2.north -| rlA2.west) -- (xlA2.south -| rlA2.west) (xlA2.north -| llA1.east) -- (xlA2.south -| llA1.east);
	\draw (xlA3.north -| rlA3.west) -- (xlA3.south -| rlA3.west) (xlA3.north -| llA2.east) -- (xlA3.south -| llA2.east);
	\draw (xlB0.north -| rlB0.west) -- (xlB0.south -| rlB0.west) (xlB0.north -| llA3.east) -- (xlB0.south -| llA3.east);
	\draw (xlB1.north -| rlB1.west) -- (xlB1.south -| rlB1.west) (xlB1.north -| llB0.east) -- (xlB1.south -| llB0.east);
	\draw (xlB2.north -| rlB2.west) -- (xlB2.south -| rlB2.west) (xlB2.north -| llB1.east) -- (xlB2.south -| llB1.east);
	\draw (xlB3.north -| rlB3.west) -- (xlB3.south -| rlB3.west) (xlB3.north -| llB2.east) -- (xlB3.south -| llB2.east);
	\draw (xlD0.north -| rlD0.west) -- (xlD0.south -| rlD0.west) (xlD0.north -| llB3.east) -- (xlD0.south -| llB3.east);
	
	\foreach \x / \y in {inpt/x000, x000/xl10, xl10/xl11, xl11/xl12, xl12/xl13, xl13/xl20, xl20/xl21, xl21/xl22, xl22/xl23, xl23/xl30, xl30/xl31, xl31/xl32, xl32/xl33, xl33/xl40,
		xl40/xl41, xl41/xl42, xl42/xl43, xl43/xl50, xl50/xl51, xl51/xl52, xl52/xl53, xl53/xl60, xl60/xl61, xl61/xl62, xl62/xl63, xl63/xl70, xl70/xl71,
		xl71/xl72, xl72/xl73, xl73/xl80, xl80/xl81, xl81/xl82, xl82/xl83, xl83/xl90, xl90/xl91, xl91/xl92, xl92/xl93, xl93/xlA0, xlA0/xlA1, xlA1/xlA2,
		xlA2/xlA3, xlA3/xlB0, xlB0/xlB1, xlB1/xlB2, xlB2/xlB3, xlB3/xlD0, xlD0/otpt}
	\draw [sqar] (\x.south) -- (\y.north);
	
	\foreach \x / \y in {xl10/xl13, xl20/xl23, xl30/xl33, xl40/xl43, xl50/xl53, xl60/xl63, xl70/xl73, xl80/xl83, xl90/xl93, xlA0/xlA3, xlB0/xlB3}
	\draw [line width=1.0pt, arrows={-{Latex[length=1.5mm, round]}}, out=-180, in=180] (\x.west) to  node [midway, fill=white, circle, draw=black, line width=0.1pt, inner ysep=0.33mm, inner xsep=0.33mm] {\texttt{\textbf{+}}} (\y.west);

	\begin{pgfonlayer}{bbackground}
	\definecolor{encC}{RGB}{205, 185, 156}
	\definecolor{repC}{RGB}{204, 204, 204}
	\definecolor{decC}{RGB}{205, 185, 156}
	\node (a) at ($(xl10.north)!0.5!(xl53.south)$) [fill=encC, draw=none, minimum width =62.5mm, minimum height=88mm, opacity=0.40] {};
	\node (b) at ($(xl60.north)!0.5!(xl63.south)$) [fill=repC, draw=none, minimum width =62.5mm, minimum height=17mm, opacity=0.50] {};
	\node (c) at ($(xl70.north)!0.5!(xlB3.south)$) [fill=decC, draw=none, minimum width =62.5mm, minimum height=88mm, opacity=0.40] {};
	\node at (a.west) [anchor=south, rotate=90] {Encoder Sub-Net};
	\node at (b.west) [anchor=south, rotate=90] {Representation};
	\node at (c.west) [anchor=south, rotate=90] {Decoder Sub-Net};
	
	\tikzstyle{CONVconcat}=[CONVst, anchor=center, minimum width=43.5mm, xshift=0.7mm, yshift=0.35mm]
	\tikzstyle{skin}=[dotted, rounded corners = 3mm, arrows={-{Latex[length=1.5mm, round]}}]
	\tikzstyle{skot}=[dashed, rounded corners = 3mm, arrows={-{Latex[length=1.5mm, round]}}]
	
	\foreach \x / \y / \m in {xl13/xlB0/17, xl23/xlA0/13, xl33/xl90/9, xl43/xl80/5, xl53/xl70/1}
	{
		\node (bnode) [CONVconcat, fill=bc01] at (\y.center) {};
		\draw [skot] (\x.east) -- 
		($(\x.east) + (3mm+\m*1.3mm, -2mm)$) -- 
		($(\y.east) + (3mm+\m*1.3mm, 2mm)$) -- 
		(bnode.east);
	}
	
	\foreach \x / \y / \m in {xl10/xlB3/20, xl11/xlB2/19, xl12/xlB1/18,
		xl20/xlA3/16, xl21/xlA2/15, xl22/xlA1/14,
		xl30/xl93/12, xl31/xl92/11, xl32/xl91/10,
		xl40/xl83/8,  xl41/xl82/7,  xl42/xl81/6,
		xl50/xl73/4,  xl51/xl72/3,  xl52/xl71/2}
	{	\node (bnode) [CONVconcat, fill=bc03] at (\y.center) {};
		\draw [skin] (\x.east) -- 
		($(\x.east) + (3mm+\m*1.3mm, -2mm)$) -- 
		($(\y.east) + (3mm+\m*1.3mm, 2mm)$) -- 
		(bnode.east);
	}
	\end{pgfonlayer}
	\end{tikzpicture}
	\caption{
		Architecture of the proposed energy monitoring model. Dashed and dotted lines to the right represent \emph{outer} and \emph{inner} skip connections, respectively. Solid lines to the left represent residual connections \cite{He_2015_ResNet}. Skip connections use channel aggregation while residual connections use element-wise addition. Green-shaded layers are followed by a pooling step, while the red-yellow shaded ones are preceded by an un-pooling operation.
	}
	\label{fig:ae-2-architecture}
\end{figure}

Figure \ref{fig:ae-2-architecture} shows the architecture of the proposed fully convolutional neural network model. The model consists of 46 layers in five parts (an input layer, 40 encoding and decoding layers, 4 representation layers, and an output layer) reaching 41M trainable parameters. Each layer includes a sequence of elementary operations shown in the figure and briefly introduced in the sequel. 

\textbf{Dilated Convolutions} \texttt{CONV(d, k)}: The core operation of each layer is the cross-correlation defined as

{\small\begin{equation*}
	\texttt{CONV(d, k):}\; f(x) \stackrel{\text{def}}{=}\;b(n) \plus \sum_{k=\minus\lfloor\texttt{k}/2\rfloor}^{k=\lfloor\texttt{k}/2\rfloor}x(n+\texttt{d}\cdot k)\cdot \kappa(k)
	\end{equation*}}%
where \texttt{d} is the dilation rate \cite{Yu_2016}, \texttt{k} is the kernel size, $\underline{b}$ is the bias vector, and $\underline{\kappa}$ is the layer's kernel.

\textbf{Batch Normalization} \texttt{BN} \cite{Ioffe_2015_BatchNormalization}: is a composition of two  affine transformations applied to the output of each layer based on mini-batch statistics

{\small\begin{equation*}
	\texttt{BN:}\;f(x) \stackrel{\text{def}}{=}\; \gamma\, \hat{x} + \beta = \gamma\, \frac{x - \mu_\mathcal{B}}{\sigma_\mathcal{B}} + \beta
	\end{equation*}}%
where $x$ is the original output of a unit, $\mu_\mathcal{B}$ and $\sigma^2_\mathcal{B}$ are the sample mean and variance of all outputs of this neuron over the mini-batch $\mathcal{B}$, and $\gamma$ and $\beta$ are two learnable parameters.

\textbf{\underline{L}eaky \underline{Re}ctified \underline{L}inear \underline{U}nits} \texttt{LReLU} \cite{Maas_2013_RectifierNonLinearities}: is a non-linear activation function defined as

{\small\begin{equation*}
	\texttt{LReLU:}\; f(x) \mathrel{\stackon[5pt]{$=$}{$\scriptscriptstyle\alpha \,\leq\,1$}} \max(\alpha x, x)
	\end{equation*}}%

\textbf{Activation noise (noise injection)} \texttt{GN} \cite{Nair_2010_ReLU_RBM}: is a regularization technique applied during the training phase only and consists of injecting small additive Gaussian noise (with variance $\sigma^2$) to the output of the layer to avoid over-fitting
\begin{equation*}
\texttt{GN:}\; f(x) \stackrel{\text{def}}{=}\; x + z \sim \mathcal{N}(0, \sigma^2)
\end{equation*}

\textbf{Sigmoidal activations} \texttt{LogSg}: is a bounded activation function applied to the first hidden layer and the output layer of the model
\begin{equation}
\texttt{LogSg:}\; f(x) \stackrel{\text{def}}{=}\; (1+\exp(-x))^{-1}
\label{eq:logsg-activation}
\end{equation}

\textbf{Down- and up-sampling:} take place only across blocks where down-sampling is performed using MaxPooling while up-sampling is applied using forward-filling. 

\textbf{Parameter initialization and updates}: model parameters are initialized from a zero-mean uniform distribution \cite{Glorot_2010} and learned using a gradient-based stochastic  optimization \cite{Bottou_2012_StochasticGradientDescent} with an update rule based on the ADAM Algorithm \cite{Kingma_2014_ADAM} with Nesterov momentum \cite{Dozat_2015_NADAM}.

\section{Performance Measures}
\label{sec:performance-measures}

Early works on energy disaggregation tended to adopt the simple accuracy index in evaluating the performance of a disaggregation system \cite{Chang_2010, Belkin_2013, Makonin_2013_AMPds1}. Later works, however, realized the misleading interpretation of this measure (resulting from its bias towards the prevailing class) and proposed \emph{precision}, \emph{recall}, and \emph{f$_1$-score} as alternative measures for assessing the disaggregation performance \cite{Beckel_2014_ECO, Holmegaard_2016_IndustrialSettings, Kim_2010, Makonin_2015_NILMPerformanceEvaluation}. 

We, however, believe that these measures represent a one-sided rigorous solution to the biasness of the accuracy index. In fact, these metrics are fused to the assumption of scarce load usage and fail to provide valuable interpretation of performance if this assumption is violated. 

Given the raw-count contingency table

\hspace{0mm}
\begin{small}
	\setlength{\textfloatsep}{0.1cm}
	\begin{tabularx}{37.5mm}{ccccc}
		&                 &       \multicolumn{2}{l}{Predictions}        & \\[1mm]
		&                 & $\hat{\omega}^\plus$ & $\hat{\omega}^\minus$ & \\[1mm]
		\cmidrule{3-4}
		\multirow{2}{0mm}{\rotatebox[origin=c]{90}{Classes}} & $\omega^\plus$  & \texttt{TP} & \texttt{FN} & \texttt{RP} \\[2mm]
		& $\omega^\minus$ & \texttt{FP} & \texttt{TN} & \texttt{RN} \\
		\cmidrule{3-4}   
		&                 & \texttt{PP} & \texttt{PN} & \texttt{N}  \\
	\end{tabularx}
\end{small}
\begin{footnotesize}
	\begin{tabularx}{10mm}{>{\hspace{0mm}}l<{\hspace{-2mm}}l<{\hspace{0cm}}}
		\texttt{TP:} & True Positives\\
		\texttt{TN:} & True Negatives\\
		\texttt{FP:} & False Positive\\
		\texttt{FN:} & False Negatives\\
		\texttt{RP:} & Real Positives \\
		\texttt{RN:} & Real Negatives\\
		\texttt{PP:} & Positive Predictions\\
		\texttt{PN:} & Negative Predictions\\
		\texttt{N:}  & Num. of samples/events
	\end{tabularx}
\end{footnotesize}\\[1mm]
the aforementioned measure are defined as

{\small
	\begin{align}
	\text{accuracy}  &= (\texttt{TP + TN})\;/\;(\texttt{TP + TN + FP + FN}) 	\label{eq:acc-TN_TP}\\
	\text{precision} &= \texttt{TPA} = \;\texttt{TP}\,\;/\;(\texttt{TP + FP})   \nonumber \\
	\text{recall}    &= \texttt{TPR} = \;\texttt{TP}\,\;/\;(\texttt{TP + FN})   \nonumber \\
	\texttt{f$_1$-s} &= (\texttt{2 x TP})\;\;/\;(\texttt{2 x TP + FN + FP})  	\label{eq:f1s-TN_TP}
	\end{align}}%
In the case of scarce load usage, the probability of negative samples becomes relatively high and the accuracy index becomes a single-sided measure, namely the true negative rate. In this case, a trivial disaggregator (one that always predicts the prevailing class) ambiguously yields near optimal accuracy.

The information retrieval approach to alleviate this bias is to simply \emph{ignore} the prevailing term in Eq. \ref{eq:acc-TN_TP}, namely \texttt{TN}, which results in either the Jaccard index or the $f$-measure \texttt{f$_1$-s} Eq. \ref{eq:f1s-TN_TP}. We find this to be an extreme and ill-argued solution, especially in assessing energy monitoring performance. First, the scarce load usage is not always valid and is usually violated in commercial buildings or some residential loads such as refrigerators, air conditioners, space heaters, or electric vehicles. Additionally, the class of always-on loads suffers from the exact opposite situation where the class imbalance is due to the prevailing positive class and a trivial system in this case yields misleading near-optimal score for both the accuracy and the $f$-measure.

Second, when the scarce usage assumption is valid (e.g. for various miscellaneous appliances such as kettles, irons, vacuum cleaners ... etc), the extent of class imbalance varies widely amongst loads as well as users. These variations are not reflected by any means in either of the information retrieval measures. For these reasons, we claimed that precision and recall are inflexible measures since they are fused to a one-sided assumption regardless of the real distribution of classes.

Powers \cite{Powers_2011_EvaluationMetrics} introduced \emph{informedness} \texttt{B}, \emph{markedness} \texttt{M}, and their geometric mean \emph{Matthews Correlation Coefficient} \texttt{MCC} as alternative, unbiased evaluation measures
\begin{align}
\texttt{B} &= \texttt{TPR + TNR - 1} \\
\texttt{M} &= \texttt{TPA + TNA - 1} \\
\texttt{MCC} &= \sqrt{\texttt{B} \cdot \texttt{M}}
\end{align}
where \texttt{TNA} = \texttt{TN / (TN + FN)} is the inverse-precision and \texttt{TNR} = \texttt{TN / (TN + FP)} is the inverse recall. Similar to the information retrieval measures, these alternatives were proposed and adopted in similar application domains such as medical diagnostics \cite{Youden_1950,Metz_1978_ROCAnalysis} and recommender system evaluations \cite{Schroder_2011_SettingGoals}. We believe that the requirements of performance evaluation in these applications are more similar to those in energy disaggregation.

\begin{table}[!t]
	\begin{center}
		\caption{Performance comparison of 11 loads from the first building in UK-DALE \cite{Kelly_2015_UKDALE}. \texttt{rn} is the probability of the negative class in the evaluation fold and \texttt{\%-NM} is the \emph{percent-noisy measure} \cite{Makonin_2015_NILMPerformanceEvaluation}.}
		\label{tbl:model-capacity}
		\setlength{\tabcolsep}{5.65pt}
		\footnotesize
		\newcommand{\fscore}{\texttt{f}$_\texttt{1}$\texttt{-s}}
		\newcommand{\wrst}[1]{\texttt{#1}}
		\newcommand{\best}[1]{{\texttt{#1}}}
		\setlength{\tabcolsep}{4.3pt}

		\begin{tabularx}{1.0\columnwidth}{c|cc|cccc|cc}
			\toprule                    
			& \texttt{rn}   & \texttt{\%-NM}  & \texttt{TPA}    & \texttt{TPR}  & \texttt{B}    & \texttt{M}    & \fscore       & \texttt{MCC} \\
			\midrule 
			\texttt{FR}             & \wrst{0.55}   & \wrst{0.87}     & \wrst{0.92}    & \wrst{0.88}  & \wrst{0.81}  & \wrst{0.82}  & \best{0.896}  & \best{0.815} \\
			\texttt{LC}             & \wrst{0.69}   & \wrst{0.92}     & \wrst{0.52}    & \wrst{0.67}  & \wrst{0.39}  & \wrst{0.35}  & \best{0.589}  & \best{0.373} \\
			\texttt{DW}             & \wrst{0.98}   & \wrst{0.95}     & \wrst{0.85}    & \wrst{0.49}  & \wrst{0.49}  & \wrst{0.84}  & \best{0.623}  & \best{0.641} \\
			\texttt{WM}             & \wrst{0.94}   & \wrst{0.91}     & \wrst{0.97}    & \wrst{0.99}  & \wrst{0.99}  & \wrst{0.96}  & \best{0.979}  & \best{0.978} \\
			\texttt{SP}             & \wrst{0.78}   & \wrst{0.97}     & \wrst{0.46}    & \wrst{0.24}  & \wrst{0.16}  & \wrst{0.27}  & \best{0.312}  & \best{0.204} \\
			\texttt{TV}             & \wrst{0.90}   & \wrst{0.97}     & \wrst{0.74}    & \wrst{0.69}  & \wrst{0.67}  & \wrst{0.71}  & \best{0.716}  & \best{0.686} \\
			\texttt{BL}             & \wrst{0.91}   & \wrst{0.95}     & \wrst{0.34}    & \wrst{0.75}  & \wrst{0.60}  & \wrst{0.31}  & \best{0.468}  & \best{0.431} \\
			\texttt{KT}             & \wrst{0.99}   & \wrst{0.95}     & \wrst{0.87}    & \wrst{0.87}  & \wrst{0.87}  & \wrst{0.87}  & \best{0.870}  & \best{0.869} \\
			\texttt{MC}             & \wrst{0.99}   & \wrst{0.97}     & \wrst{0.62}    & \wrst{0.46}  & \wrst{0.45}  & \wrst{0.62}  & \best{0.526}  & \best{0.529} \\
			\texttt{TS}             & \wrst{0.99}   & \wrst{0.98}     & \wrst{0.67}    & \wrst{0.72}  & \wrst{0.72}  & \wrst{0.67}  & \best{0.698}  & \best{0.697} \\
			\texttt{KL}             & \wrst{0.87}   & \wrst{0.94}     & \wrst{0.46}    & \wrst{0.55}  & \wrst{0.46}  & \wrst{0.39}  & \best{0.502}  & \best{0.422} \\
			\bottomrule
		\end{tabularx}

	\end{center}
\end{table}

\begin{table}[!t]
	\begin{center}
		\caption{Performance comparison between the proposed model \texttt{AE} and the \emph{rectangles} architecture in \cite{Kelly_2015} \texttt{Regr} on same load instances (left) and unseen instances from new buildings (right). All values represent the $f$-measure.}
		\label{tbl:across-loads}
		\setlength\intextsep{0mm}
		\setlength{\tabcolsep}{10pt}
		\footnotesize
		\newcommand{\fscore}{\texttt{f}$_\texttt{1}$\texttt{-s}}
		\newcommand{\wrst}[1]{\texttt{#1}}
		\newcommand{\best}[1]{\textbf{\texttt{#1}}}
		\begin{tabularx}{0.94\columnwidth}{c|cc|cc}
			\toprule					
			\multirow{2}{*}{Load}	&				\multicolumn{2}{c}{Same instances} 					&		\multicolumn{2}{c}{Accross buildings} 				\\
			\cmidrule(lr){2-3}
			\cmidrule(lr){4-5}
			&	\texttt{Regr.} \cite{Kelly_2015}	&		\texttt{AE}		&		\texttt{Regr.} \cite{Kelly_2015}		& \texttt{AE}	\\
			\midrule
			\texttt{FR}				&		\wrst{0.810} 					&		\best{0.879}		&			\wrst{0.820} 					& \best{0.927}	\\
			\texttt{DW}				&		\wrst{0.720} 					&		\best{0.796}		&			\wrst{0.740} 					& \best{0.804}	\\
			\texttt{MC}				&		\wrst{0.620} 					&		\best{0.705}		&			\wrst{0.210} 					& \best{0.366}	\\
			\texttt{WM}				&		\wrst{0.490} 					&		\best{0.960}		&			\wrst{0.270} 					& \best{0.410}	\\
			\texttt{KT}				&		\wrst{0.710} 					&		\best{0.783}		&			\wrst{0.700} 					& \best{0.819}	\\
			\bottomrule
		\end{tabularx}
	\end{center}
\end{table}

\section{Experiments and Results}
\label{sec:experiments-and-results}

The developed model is evaluated on the freely available UK-DALE dataset \cite{Kelly_2015_UKDALE}, an energy dataset acquired from five residential buildings.
In this work, the 1 Hz real power measurements represent the input signals to disaggregate while the reference ones are the 1/6 Hz measurements up-sampled (using fill-forward) to 1 Hz.

Table \ref{tbl:model-capacity} shows the performance measures of the  proposed model evaluated on 11 loads from the first building in the adopted dataset with a 3-hour monitoring window for all load categories. Data folds are real power measurements from January and February of 2015 for training and validation, respectively, while the remaining 10 months of the 2015 represents the evaluation fold. While we provide these results as benchmarking ones, assessment of feasibility is observed in the following experiment.

In Table \ref{tbl:across-loads}, we compare the monitoring performance of our model \texttt{AE} with the previous work in \cite{Kelly_2015}, specifically the regression-based model \texttt{Regr.} (referred to as \emph{rectangles} architecture). We use the exact data folds adopted in \cite{Kelly_2015} for training and evaluation and define two test cases. The first trains and evaluates on the same load instances but future periods of operation while the second evaluates on new load instances (from new buildings). In both cases, the proposed model outperformed previous works in all load categories.

\section{Conclusion and future work}
\label{sec:conclusion}

In this paper, we assessed the feasibility of a generic deep disaggregation model for end-use load monitoring using a fully convolutional neural network evaluated on a variety of load categories.
The proposed model (with a fixed architecture and set of hyper-parameters) outperforms previous work and showed relatively acceptable performance across different loads.

{\small\vspace{-1mm}
\bibliographystyle{./IEEEtranKarim}
\bibliography{./IEEEabrv,./References}}

\end{document}